\documentclass{article}

\usepackage{microtype}
\usepackage{graphicx}
\usepackage{subfigure}
\usepackage{booktabs} 

\usepackage{hyperref}

\usepackage[accepted]{icml2024}

\usepackage{amsmath}
\usepackage{amssymb}
\usepackage{mathtools}
\usepackage{amsthm}

\usepackage[capitalize,noabbrev]{cleveref}

\usepackage{float}
\usepackage{acronym}
\usepackage{listings}
\usepackage{subcaption}
\usepackage{array}
\usepackage{svg}
\usepackage{tikz}
\usepackage{svg}
\usepackage{ulem}
\usetikzlibrary{matrix,calc}

\usepackage{fontawesome} 
\usepackage{bm}
\usepackage{caption}
\theoremstyle{plain}

\theoremstyle{definition}

\theoremstyle{remark}

\usepackage[textsize=tiny]{todonotes}

\icmltitlerunning{DoubleML\textsubscript{Deep}}

\begin{document}

\twocolumn[
\icmltitle{DoubleML\textsubscript{Deep}: Estimation of Causal Effects with Multimodal Data}



\icmlsetsymbol{equal}{*}

\begin{icmlauthorlist}
\icmlauthor{Sven Klaassen}{comp,UHH}
\icmlauthor{Jan Teichert-Kluge}{UHH}
\icmlauthor{Philipp Bach}{UHH}
\icmlauthor{Victor Chernozhukov}{MIT}
\icmlauthor{Martin Spindler}{comp,UHH}
\icmlauthor{Suhas Vijaykumar}{MIT}
\end{icmlauthorlist}

\icmlaffiliation{UHH}{University of Hamburg, Germany}
\icmlaffiliation{comp}{Economic AI GmbH, Germany}
\icmlaffiliation{MIT}{Massachusetts Institute of Technology, USA}

\icmlcorrespondingauthor{Sven Klaassen}{sven.klaasen@uni-hamburg.de}

\icmlkeywords{Causal Machine Learning, Double Machine Learning, Causal Inference, Deep Learning}

\vskip 0.3in
]



\printAffiliationsAndNotice{\icmlEqualContribution} 

\begin{abstract}
This paper explores the use of unstructured, multimodal data, namely text and images, in causal inference and treatment effect estimation.
We propose a neural network architecture that is adapted to the double machine learning (DML) framework, specifically the partially linear model.
An additional contribution of our paper is a new method to generate a semi-synthetic dataset which can be used to evaluate the performance of causal effect estimation in the presence of text and images as confounders.
The proposed methods and architectures are evaluated on the semi-synthetic dataset and compared to standard approaches, highlighting the potential benefit of using text and images directly in causal studies. Our findings have implications for researchers and practitioners in economics, marketing, finance, medicine and data science in general who are interested in estimating causal quantities using non-traditional data.
\end{abstract}

\section{Introduction}
In this paper, we delve into the realm of causal inference and treatment effect estimation in the presence of high-dimensional and unstructured multimodal confounders, emphasizing the utilization of deep learning techniques for handling complex nuisance parameters. In many cases, text and image data contain information that can otherwise not be accounted for in causal studies, for example in the form of sentiment in product descriptions or reviews, labels for product images in online marketplaces or health information encoded in medical images. In causal studies, this information can be very important to account for otherwise unmeasured confounding or to improve estimation precision of causal effects.
Our focus is on developing methods that ensure root-$N$ consistency and valid inferential statements of the causal parameter, particularly in scenarios where traditional semi-parametric assumptions are challenged by the increasing complexity of the nuisance parameter space \citep{Foster2023}. The parameter of interest will typically be a causal or treatment effect parameter, denoted by $\theta_0$. Common examples for $\theta_0$ include the average treatment effect (ATE) or the ATE for the subgroup of the Treated (ATT). We consider settings in which the nuisance parameters / functions will be estimated using deep learning methods, such as transformers, or large language models (LLM). These deep learning methods are capable of handling high-dimensional, unstructured covariates, like texts and images \citep{Goodfellow2016, zhang2023dive}, and provide estimators of nuisance functions when these functions are highly complex. In this context, "highly complex" means that the entropy of the parameter space associated with the nuisance parameter increases with the sample size, going beyond the conventional framework addressed in the classical semi-parametric literature \citep{härdle2012nonparametric}. The main contribution of this paper is to offer a general procedure for estimation and inference on $\theta_0$ that is formally valid in these highly complex settings. In Section \ref{synth_data}, we also propose a method to generate semi-synthetic data in the presence of text and images as confounders. Data generating processes for unstructured data are characterized by inherent challenges, which are briefly summarized and addressed in our paper. Given the growing interest in causal inference with text and image data and the increased availability of this data, we believe that this contribution of our paper might be of independent interest.\\
As a lead example, we consider the following partially linear regression (PLR) model \citep{chernozhukov2018}:
\begin{align}
Y &= \theta_0 D + g_0(X) + \varepsilon, & \mathbb{E}[\varepsilon | X, D] = 0 \label{eq:plr1} \\
D &= m_0(X) + \vartheta, & \mathbb{E}[\vartheta | X] = 0 \label{eq:plr2}
\end{align}
Here, $Y$ is the outcome variable, $D$ is the policy/treatment variable of interest,  $X$ consists of other controls, and $\varepsilon$ and $\vartheta$ are disturbances. The first equation is the main equation, and $\theta_0$ is the main regression coefficient we would like to infer. If $D$ is conditionally exogenous on controls $X$, $\theta_0$ has the interpretation of the average treatment effect parameter or the "lift" parameter in business applications. The second equation keeps track of confounding, namely the dependence of the treatment variable on controls. This equation is not of interest per se but is important for characterizing and removing regularization bias. Confounders $X$ affect the policy variable $D$ through the function $m_0(X)$ and the outcome variable via the function $g_0(X)$. Not correctly accounting for all confounding factors, e.g. by not including all relevant confounders $X$, may lead to biased estimates of the target parameter $\theta_0$. In real world applications the confounding factors might be very complex and hard to observe or measure (e.g. product quality or medical information). Text and image data can be helpful to control for these complex confounding factors, for instance product images and descriptions are usually a good indicator of product quality. Consequently, causal models can benefit from text and image data to remove selection bias. We leverage deep learning methods to fit functions representing the conditional expectations of the output variable $Y$ and the treatment variable $D$ given our set of covariates. Specifically, we define:
\begin{align}
l_0(X) &:= \mathbb{E}[Y | X] \label{eq:l_hat} \\
m_0(X) &:= \mathbb{E}[D | X] \label{eq:m_hat}
\end{align}
To construct an orthogonalized score for the fitted nuisance learners $\hat{\eta} = (\hat{l}, \hat{m})$, we define the following expression:
\begin{align*}
\begin{split}
\psi(W, \theta, \hat{\eta}) := &\left(Y - \hat{l}(X) - \theta\left(D - \hat{m}(X)\right)\right) \\ &\cdot \left(D - \hat{m}(X)\right),
\end{split}
\end{align*}
where $W=(Y,D,X)$. The construction of an orthogonalized score ensures the necessary orthogonality for valid causal inference. Finally, the estimate is computed as the solution to the following equation:
\begin{align*}
0 = \frac{1}{n}\sum_{i=1}^n \psi(W, \hat{\theta}, \hat{\eta}_0)
\end{align*}
The assumptions involve bounding the difference between estimated nuisance functions (\(\hat{m}\) and \(\hat{l}\)) and true functions (\(m_0\) and \(\ell_0\)) in relation to the sample size $N$
\begin{align}
\begin{split}
&\lVert \hat{m}(X) - m_0(X) \rVert_{P,2} \\ &\cdot \left(\lVert \hat{m}(X) - m_0(X) \rVert_{P,2} +  \lVert \hat{l}(X) - l_0(X)\rVert _{P,2}\right) \\ \le & \delta_N N^{-1/2}.
\end{split}
\label{eq:comb_loss}
\end{align}
Under this assumption and additional regularity conditions\footnote{$\delta_N$ is a sequence of positive constants converging to zero. For details on the regularity conditions, see \citet{chernozhukov2018}.}, the estimator \(\hat{\theta}\) converges to the true parameter \(\theta_0\) at a rate of \(1/\sqrt{n}\) and is approximately normally distributed:
\begin{align*}
\sqrt{n}(\hat{\theta} - \theta_0) \to \mathcal{N}(0, \sigma^2)
\end{align*}
This methodology forms the basis for treatment effect estimation in the presence of high-dimensional unstructured confounders.
In modern research, the availability of unstructured data, such as images and text, has become ubiquitous. These data types offer a rich source of information that can contribute significantly to the estimation of causal effects. Incorporating unstructured data into the causal models such as the PLR as controls introduces several advantages. First, it allows for a more comprehensive representation of the confounding structure, capturing nuances that may be missed by solely relying on structured / tabular covariates. Second, more and more deep learning models are being developed which are tailored for unstructured data, such as transformers for images or LLMs for text, can be seamlessly integrated into the estimation process, further enhancing the accuracy of nuisance parameter estimation \citep{chernozhukov2018}. Third, we would like to note that we focus specifically on the semi-parametric PLR model in this paper. In general, our methodology is basically also applicable to other nonparametric causal models that share the key ingredients of the double machine learning framework (cf. Section \ref{dml_bsl}). \\ \\
In the subsequent sections, we elaborate on the methodology for leveraging unstructured data within the PLR framework and present simulation studies based on the semi-synthetic dataset.

\section{Literature Review and Examples}
The use of unstructured data such as images and text as controls is crucial for identification and estimation of causal parameters, for example for estimating price elasticities, effects of medical treatments \citep{chang2018, masukawa2022}, or the effect of condensation trails on the climate \citep{wu2023}. While unstructured data have been used for prediction tasks for some time, e.g. deep learning techniques for clinical risk predictions \citep{zhang2020}, the use of text and images for causal inference has been a very recent development in the scientific literature. 
Text as outcome and treatment variable was discussed in \citet{egami2018how} and \citet{Blei2022}, but on a very high / conceptual level. We focus on text (and images) as confounders. \citet{Veitch20a} consider this setting and provide results for the consistency of the causal estimate, while we integrate it into the double machine learning framework to perform valid inference, i.e. constructing valid confidence intervals and test statistics for causal parameters. \citet{causaltrans} apply transformer to estimate treatment effects with tabular data and time-varying covariates, but also not provide inference results. The recent literature on the use of text for causal inference is nicely summarized in \citet{Feder2022}. There are also approaches to use images in causal inference \citep{Jerzak2023, Jerzak2023imagebased, Jerzak2023integrating}, but we are, to the best of our knowledge, not aware of any study allowing for valid inference with images and consider our approach as the first to integrate both text and images in a \textit{multimodal} double machine learning framework.\\

\section{Getting started / Warm up: Double Machine Learning for Tabular Data} \label{dml_bsl}
The double machine learning method, as proposed by \citet{chernozhukov2018}, is a framework that aims to provide valid statistical inference in structural equation models while leveraging the predictive power of machine learning methods for potentially high-dimensional and non-linear nuisance functions. The DML framework consists of three key ingredients:
\begin{itemize}
    \item Neyman orthogonality
    \item High-quality machine learning estimation
    \item Sample splitting
\end{itemize}
Neyman orthogonality ensures that the score function $\psi(W, \theta, \hat{\eta})$ to estimate the target parameter $\theta_0$ is insensitive towards the plug-in estimates from the nuisance learners $\hat{\eta}$. This is specifically relevant for machine learning algorithms such as neural networks, as these usually trade of variance and bias to achieve high-quality predictions via regularization.\\
High-quality machine learning estimation involves using state-of-the-art machine learning algorithms to estimate the nuisance functions in Equations \ref{eq:l_hat} and \ref{eq:m_hat}, ensuring that they are estimated as accurately as possible.\\
Sample splitting is employed to separate the data into estimation and inference samples, which helps to avoid overfitting, and hence to achieve valid statistical inference.

The DML framework has gained attention in various domains, including social sciences, computer science, medicine, biostatistics, and economics and finance, because of its ability to address the challenges of causal inference with high-dimensional data.
Furthermore, the DML framework has been implemented in both R and Python programming languages as \texttt{DoubleML} \cite{bach2021}, making it accessible to many users in academic and industry research.
In summary, the double machine learning method, as described by \citet{chernozhukov2018}, provides a robust framework for valid statistical inference for tabular data or structured data in structural equation models by integrating high-quality machine learning estimation with sample splitting. Its versatility and applicability across various domains make it a valuable tool for addressing causal inference challenges with complex, high-dimensional controls. 

\section{Double Machine Learning for Text and Images}
In the simplest case, the covariates $X$ are tabular, as discussed in the previous section, and can be represented as $X_{\text{tab}} = (X_1, \ldots, X_p)$ such that $X_{\text{tab}} \in \mathbb{R}^p, \,\, p \in \mathbb{N}$. However, if the controls are unstructured, e.g. in the simple form of RGB images, $X$ can be represented as a tensor such that $X_{\text{img}} \in \mathbb{R}^{3 \times h \times w}, \,\, h, w \in \mathbb{N}$.
For controls in text form, $X$ can be defined as an input representation matrix including the token, segmentation and position embeddings \citep{devlin2019bert} and can therefore be written as $X_{\text{txt}} \in \mathbb{R}^{3 \times S}, \,\, S \in \mathbb{N}$ where $S$ denotes the sequence length of the input sentence.

\begin{figure}
    \centering
    \subfigure[]{
        \label{fig:DAG_txt_img_subfig1}
        \begin{tikzpicture}
            \tikzset{> = stealth,
            circ_targets/.style = {
                draw = black,
                fill=lightgray,
                shape = circle,
                inner sep = 1pt
            },
            circ/.style = {
                draw = black,
                shape = circle,
                inner sep = 1pt
            }}
            \node[circ_targets] (D) at (0,0) {\bm{$D$}};
            \node[circ_targets] (y) at (2,0) {\bm{$Y$}};
            \node[circ, align=center] (features) at (1,1.2) {\faFileText\quad \faImage \\ \faTable};
            \path[->, black] (features) edge (D);
            \path[->, black] (features) edge (y);
            \path[->, black, line width=1.5pt] (D) edge (y);
            \draw (D) -- (y) node [midway, fill=white] {$\theta_0$};
        \end{tikzpicture}
    }
    \hspace{1em} 
    \subfigure[]{
        \label{fig:DAG_txt_img_subfig2}
        \begin{tikzpicture}
            \tikzset{> = stealth,
            circ_targets/.style = {
                draw = black,
                fill=lightgray,
                shape = circle,
                inner sep = 1pt
            },
            circ/.style = {
                draw = black,
                shape = circle,
                inner sep = 1pt
            }}
            \node[circ_targets] (D) at (0,0) {\bm{$D$}};
            \node[circ_targets] (y) at (2,0) {\bm{$Y$}};
            \node[circ] (features_X) at (2,1) {\bm{$X$}};
            \node[black] (txt) at (1,0.75) {\faFileText};
            \node[black] (img) at (1,1.5) {\faImage};
            \path[->, black] (features_X) edge (txt);
            \path[->, black] (features_X) edge (img);
            \path[->, black] (txt) edge (D);
            \path[->, black] (img) edge (D);
            \path[->, black] (features_X) edge (y);
            \path[->, black, line width=1.5pt] (D) edge (y);
            \draw (D) -- (y) node [midway, fill=white] {$\theta_0$};
        \end{tikzpicture}
    }
    \caption{Examples of directed acyclic graphs (DAGs) with image and text confounding. (a) Direct confounding via image, text and tabular data. (b) Treatment decision is driven by text and images. All backdoor paths are blocked by conditioning on both image and text data.}
    \label{fig:DAG_txt_img}
\end{figure}
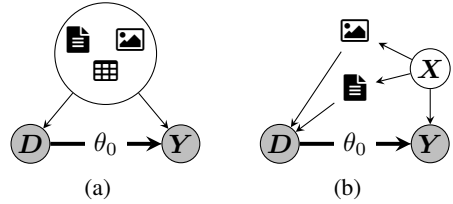

If all input modalities are to be used together as controls, $X$ can be represented as a set consisting of $X = (X_{\text{tab}}, X_{\text{txt}}, X_{\text{img}})$ with $X_{\text{tab}} \in \mathbb{R}^p$ for the tabular input, $X_{\text{txt}} \in \mathbb{R}^{3 \times S}$ for the text input and $X_{\text{img}} \in \mathbb{R}^{3 \times h \times w}$ for the image input with $p, h, w, S \in \mathbb{N}$. This results in a high-dimensional input vector, which cannot be directly used for nuisance estimation, but rather can be modeled as input for Deep Learning architectures, resulting in low-dimensional representations.
The influence of text and image data is illustrated in Figure \ref{fig:DAG_txt_img}.

As highlighted in the previous sections, the DML approach relies on high-quality estimates for the nuisance elements $m_0(X)$ and $l_0(X)$, such that the mean squared error (or product of root mean squared errors in Equation \ref{eq:comb_loss}) converges fast enough. For tabular data these rates are achievable via standard machine learning algorithms such as e.g. lasso \citep{BRT2009} or boosting \citep{luo2022highdimensional}. Theoretical results on the convergence rates of neural networks are not that clear, due to the large difference in architectures. Among others, see \citet{schmidt2020nonparametric}, \citet{kohler2021rate} or \citet{farrell2021deep}. These results are mostly geared towards feed forward networks, but highlight that the required theoretical rates are achievable.
Nevertheless, more complex neural network architectures as e.g. transformers achieve stunning predictive performance in the respective regression or classification tasks, suggesting credibly fast convergence and therefore likely fulfilling the conditions for the double machine learning framework.

\subsection{Model}
To estimate the nuisance functions according to Equations \ref{eq:l_hat} and \ref{eq:m_hat} with the set of multimodal controls $X$, we need advanced methods. The focus of this work is on multimodal models due to their promising results across different tasks. Utilizing multimodal data fusion is particularly suited for achieving the objective of estimating the causal parameter $\theta_0$ incorporating confounders sourced from tabular features, text and images. Multimodal models are a class of ML models that can effectively handle input data from different modalities such as text, image, video or audio. These models combine information from multiple modalities to improve their predictive power and achieve a better performance compared to individual modalities systems \citep{rahate2022multimodal}. Multimodal data fusion seeks to extract and merge contextual information from multiple modalities in order to enhance decision-making. This is done by taking advantage of the complementary strengths of each modality \citep{lipkova2022artificial}. Multimodal models can, for example, combine semantic knowledge gained from texts with knowledge of spatial structures obtained from images to learn joint representations of images and texts \citep{miller2021multimodal}. The objective of a multimodal model is to combine features of various modalities \citep{lee2022multimodal}. The architecture of these models can be diverse and includes, for example, neural networks capable of processing and analyzing each modality, followed by a fusion module that concatenates the information across modalities \citep{miller2021multimodal}.

\subsection{Deep Learning Architecture and Implementation Details}
\begin{figure}[]
    \centering
    \centerline{\includegraphics[width=\columnwidth]{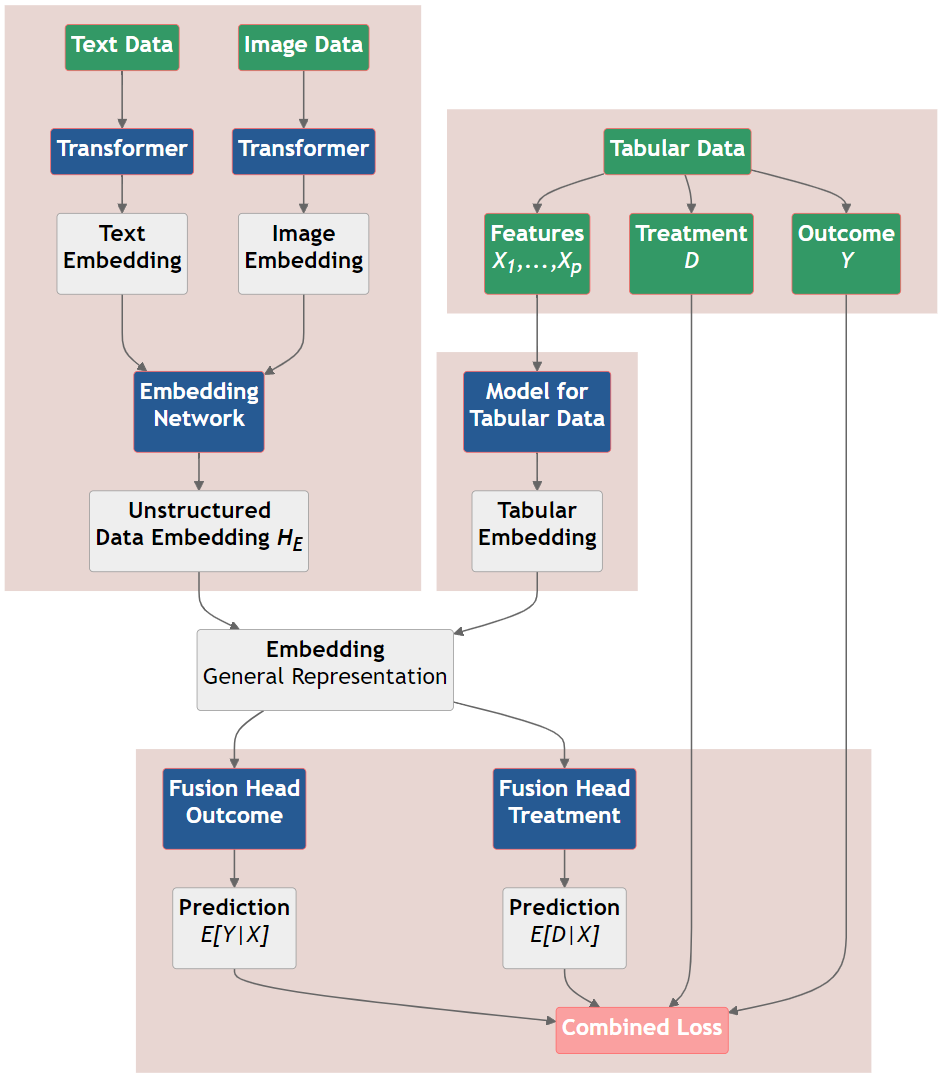}}
    \caption{High-Level PLR Model Architecture. Both nuisance components are trained simultaneously with a combined loss.}
    \label{fig:PLRNetwork}
\end{figure}

The high-level architecture of our model for the single treatment case is shown in Figure \ref{fig:PLRNetwork}. The integration of the modalities occurs through a middle fusion approach, whereby the text and image data are combined at an intermediate representation level, utilizing the embedding output. This hidden state is processed by a linear layer and an activation function. Finally, the embedding $H_E \in \mathbb{R}^E$ provides a $E$-dimensional representation of the input data that can be used to make predictions or classify the input \citep{wolf-etal-2020-transformers}.\\
We use pre-trained transformer-based models such as BERT \citep{vaswani2023attention} or variants like the robustly optimized BERT Pretraining Approach (RoBERTa) \citep{liu2019roberta} or LLMs like Llama \citep{touvron2023llama} for handling the text data. For the image processing, we rely on transformer-based models such as BEIT \citep{bao2022beit} or Vision Transformers (VITs) \citep{dosovitskiy2021image}. Using models like the SAINT model \citep{somepalli2021saint} for tabular data appears to be satisfactory. 

It is crucial to closely monitor losses of each nuisance component to ensure high-quality predictions:
\begin{align*}
    \lVert Y - \hat{l}(X) \rVert_{P,2} &\le \lVert l_0 - \hat{l}(X) \rVert_{P,2} + \sigma_{\epsilon} \\
    \lVert D - \hat{m}(X) \rVert_{P,2} &\le \lVert m_0(X) - \hat{m}(X)\rVert_{P,2} + \sigma_{\vartheta}.
\end{align*}
These considerations contribute to the robustness and effectiveness of the models applied to both unstructured and tabular data, emphasizing the importance of vigilance over nuisance losses. The combined loss function for model training is defined as the product of root mean squared errors:
\begin{align*}
    L = \lVert D - \hat{m}(X) \rVert_{P_n,2}\lVert Y - \hat{l}(X)  \rVert_{P_n,2}.
\end{align*}

Usually the DML framework relies on cross-fitting to ensure the nuisance learners do not overfit. In principle cross-fitting is also possible for neural networks but would increase the computational burden excessively. Instead, we rely on simple sample splitting, such that the neural network is trained on a training set and the estimation of the target parameter $\theta_0$ is performed on a seperate test set. Recent simulation results by \citet{dmlhyper} suggest that the performance of DML with a single sample split is comparable to a cross-fitting approach in large sample sizes.

\section{Simulation Study}
The validation of the proposed estimator is a crucial step in ensuring their accuracy and reliability. Unlike predictive models, the performance of causal models is not straightforward to evaluate in real world applications. The evaluation of causal estimation approaches is generally complicated by the fact that the true causal effect (unlike true labels for predicted outcomes) is not observable, which requires the use of synthetic or semi-synthetic data. In this simulation study, we generate a semi-synthetic dataset with a known treatment effect parameter. We document a generally inherent challenge of simulating multimodal data for causal estimation: Generating credible confounding through unstructured data makes it very hard to uncover the true causal effect parameter. In our data generating process, the confounding operates through labels of the supervised learning tasks such as image classification, which cannot be perfectly predicted by the neural nets. Consequently, a part of the imposed confounding remains unexplained, which prevents exact identification and estimation of the causal parameter. Hence, we consider the true causal parameter as an ideal, but generally infeasible statistical estimate.
Our analysis will include simulations to evaluate prediction performance for treatment and outcome, joint loss function values, and variance and bias of the causal effect estimate.

\subsection{Simulating Confounding with Text and Images} \label{synth_data}
To evaluate the performance of our model, we generate a semi-synthetic dataset according to the underlying model in Equations \ref{eq:plr1} and \ref{eq:plr2}
\begin{align*}
Y &= \theta_0 D + \tilde{g}_0(\tilde{X}) + \varepsilon, \\
D &= \tilde{m}_0(\tilde{X}) + \vartheta, 
\end{align*}
where $\tilde{X}= (\tilde{X}_{\text{tab}}, \tilde{X}_{\text{txt}}, \tilde{X}_{\text{img}})$ with the following additive structure
\begin{align*}
\tilde{g}_0(\tilde{X}) &= \sum_{\text{mod}\in\{\text{tab},\text{txt},\text{img}\}} \tilde{g}_{\text{mod}}(\tilde{X}_{\text{mod}}) \\
\tilde{m}_0(\tilde{X}) &= \sum_{\text{mod}\in\{\text{tab},\text{txt},\text{img}\}} \tilde{m}_{\text{mod}}(\tilde{X}_{\text{mod}}) 
\end{align*}
and $\varepsilon, \vartheta \sim \mathcal{N}(0, 1)$.\\
Each of the three modality effects is based on a publicly available (simple) non-synthetic dataset which is usually used for classification and regression tasks. All datasets contain one target $\tilde{X}_{\text{mod}}$ (outcome or label) and features $X_{\text{mod}}$ (image, text etc.). As these datasets have been shown to work well with the respective predictive task, we generate the confounded treatment $D$ and outcome $Y$ based on the targets $\tilde{X}_{\text{mod}}$ of the three different datasets instead of the respective features $X_{\text{mod}}$. This ensures a credible confounding, especially for image and text data as the confounding effect depends on content of the image features. Further, the dataset is close to non-synthetic data while still adhering to the partially linear model.

The tabular data is sourced from the DIAMONDS dataset \citep{ggplot2_book}, which includes various attributes of diamonds such as carat, cut, color, clarity, depth, table, price, and measurements (x, y, z). The logarithm of the price column $\tilde{X}_{\text{tab}}$ is used to simulate confounding and the other variables will be the tabular input $X_{\text{tab}}$. The tabular data is preprocessed (log-transformation etc.) and downsampled to create a dataset with $N = 50,000$ observations to maintain consistency with the other data modalities of the new semi-synthetic dataset.\\
The text data component of the semi-synthetic dataset is derived from the IMDB dataset \citep{maas-EtAl:2011:ACL-HLT2011}, which is a collection of movie reviews with corresponding sentiment labels. This dataset is publicly available and has been widely used for sentiment analysis tasks in natural language processing research. The semi-synthetic dataset is based on the training and test sample of the IMDB dataset to match the size of the tabular and image datasets. The binary representation of the sentiment $\tilde{X}_{\text{txt}}$ is used to generate confounding while the review constitutes the text input $X_{\text{txt}}$ for the set of controls.\\
The image data component of the semi-synthetic dataset is sourced from the CIFAR-10 dataset \citep{Krizhevsky09learningmultiple}, which is a well-known benchmark in the field of computer vision. The CIFAR-10 dataset consists of 60,000 32x32 color images in 10 different classes, with 6,000 images per class. The semi-synthetic dataset is based on the training set, which contains 50,000 images. A numerical representation of the image labels $\tilde{X}_{\text{img}}$ is used to obtain a confounding on $Y$ and $D$ while the images will be part of the set of controls as $X_{\text{img}}$.

The effect on the outcome $Y$ is generated via a standardized version of target variable
\begin{align*}
    \tilde{g}_{\text{mod}}(\tilde{X}_{\text{mod}}) = \frac{\tilde{X}_{\text{mod}} - \mathbb{E}[\tilde{X}_{\text{mod}}]}{\sigma_{\tilde{X}_{\text{mod}}}}, \quad \text{mod}\in\{\text{tab}, \text{txt}, \text{img}\}
\end{align*}
to balance the confounding impact of all modalities. Further, the impact on the treatment $D$ is defined via
\begin{align*}
    \tilde{m}_{\text{mod}}(\tilde{X}_{\text{mod}}) = -\tilde{g}_{\text{mod}}(\tilde{X}_{\text{mod}}), \quad \text{mod}\in\{\text{tab}, \text{txt}, \text{img}\}
\end{align*}
to ensure a strong confounding. Due to the negative sign and the additive structure, the confounding effect will ensure that higher outcomes $Y$ occur with lower treatment values $D$, creating a negative bias. Further, the independence of all three original datasets and the additive negative confounding results in a negative bias even if we only control for a subset of confounding factors.\\
The treatment effect is set to $\theta_0=0.5$ and both $\tilde{g}_0(X)$ and $\tilde{m}_0(X)$ are rescaled to ensure a signal-to-noise ratio of $2$ for $Y$ and $D$ (given unit variances of the error terms).
\begin{figure}
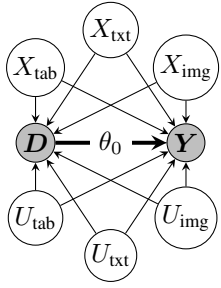

    \centering
    \tikzset{> = stealth,
    circ_targets/.style = {
        draw = black,
        fill=lightgray,
        shape = circle,
        inner sep = 1pt
    },
    circ/.style = {
        draw = black,
        shape = circle,
        inner sep = 1pt
    }}
    \tikz{
        \node[circ_targets] (D) at (0,0) {\bm{$D$}};
        \node[circ_targets] (y) at (2,0) {\bm{$Y$}};
        \node[circ] (x_tab) at (0,1) {$X_{\text{tab}}$};
        \node[circ] (x_txt) at (1,1.5) {$X_{\text{txt}}$};
        \node[circ] (x_img) at (2,1) {$X_{\text{img}}$};
        \node[circ] (U_tab) at (0,-1) {$U_{\text{tab}}$};
        \node[circ] (U_txt) at (1,-1.5) {$U_{\text{txt}}$};
        \node[circ] (U_img) at (2,-1) {$U_{\text{img}}$};
        \path[->, black] (x_tab) edge (D);
        \path[->, black] (x_tab) edge (y);
        \path[->, black] (U_tab) edge (D);
        \path[->, black] (U_tab) edge (y);
        \path[->, black] (x_txt) edge (D);
        \path[->, black] (x_txt) edge (y);
        \path[->, black] (U_txt) edge (D);
        \path[->, black] (U_txt) edge (y);
        \path[->, black] (x_img) edge (D);
        \path[->, black] (x_img) edge (y);
        \path[->, black] (U_img) edge (D);
        \path[->, black] (U_img) edge (y);
        \path[->, black, line width=1.5pt] (D) edge (y);
        \draw (D) -- (y) node [midway, fill=white] {$\theta_0$};
    }
    \caption{DAG for the semi-synthetic dataset. The confounding via the features $X=(X_{\text{tab}}, X_{\text{txt}}, X_{\text{img}})$ can be adjusted for, whereas the unexplained/noise parts $U=(U_{\text{tab}}, U_{\text{txt}}, U_{\text{img}})$ are unobserved.}
    \label{fig:conf_dag}
\end{figure}

A generally inherent challenge of this type of data generating processes is the dependency on the target of the modality $\tilde{X}_{\text{mod}}$, which might not be fully explained by the corresponding features $X_{\text{mod}}$. For example, the price of the DIAMONDS dataset can not be perfectly predicted, introducing a small part of confounding which can not be controlled for by using the tabular features $X_{\text{tab}}$ instead of $\tilde{X}_{\text{tab}}$ (logarithm of price), as shown in the DAG in Figure \ref{fig:conf_dag}. Consequently, the estimate $\hat{\theta}$ might only be able to account for the part of confounding which can be explained by the input features as
\begin{align*}
    \tilde{X}_{\text{mod}} = \mathbb{E}[\tilde{X}_{\text{mod}}|X_{\text{mod}}] + U_{\text{mod}},
\end{align*}
where $U_{\text{mod}}$ can not be controlled for.
Nevertheless, since all modalities contribute a negative bias, the semi-synthetic dataset can be used as a benchmark with an oracle upper bound of an effect estimate of $\theta_0=0.5$. To evaluate the confounding one can evaluate a basic ordinary least squares model with outcome $Y$ on the treatment variable $D$ (excluding all confounding variables). The resulting effect estimate
$$\hat{\theta}_{\text{OLS}} = -0.4594,$$
can be interpreted as a lower bound for the effect estimate.
Accordingly, all evaluated models should estimate the parameter between $-0.4594$ and $0.5$, where higher values indicate better bias correction (ignoring sampling uncertainty).\\
To further access the predictive performance of the nuisance models, we can rely on oracle predictions of
\begin{align*}
    \tilde{m}_0(\tilde{X}):= \mathbb{E}[D|\tilde{X}] \\
    \tilde{l}_0(\tilde{X}):= \mathbb{E}[Y|\tilde{X}] &= \theta_0 \tilde{m}_0(\tilde{X}) + \tilde{g}_0(\tilde{X}).
\end{align*}
Evaluating the oracle predictions $\tilde{m}_0(\tilde{X})$ and $\tilde{l}_0(\tilde{X})$ results in the following upper bounds for the performance of the nuisance estimators
\begin{small}
\begin{align*}
    R^2(D, \tilde{m}_0(\tilde{X})) &= 0.6713\\
    R^2(Y, \tilde{l}_0(\tilde{X})) &= 0.5845
\end{align*}
\end{small}
on the whole dataset of $N=50,000$ observations, which is to be expected due to the choice of signal-to-noise ratio. Again, since models only have access to the features $X=(X_{\text{tab}}, X_{\text{txt}},  X_{\text{img}})$ instead of the targets $\tilde{X}=(\tilde{X}_{\text{tab}}, \tilde{X}_{\text{txt}},  \tilde{X}_{\text{img}})$, the above values represent lower and upper bounds for $R^2$.

\subsection{Results} \label{results}
In this section, two different estimation approaches based on the proposed architecture are evaluated on the semi-synthetic dataset and compared to a baseline model.
To ensure that all models are comparable and only differ correction for confounding, we rely on the implementation of the partially linear regression model from the \texttt{DoubleML} package \citep{bach2022doubleml}. All models only differ in the nuisance estimates, which are passed to the \texttt{DoubleML} implementation.

The \textit{Baseline Model} is a standard DML approach, relying only on tabular data. The estimation of the nuisance elements is based on the \texttt{LightGBM} package \citep{NIPS2017_6449f44a} only using the features $X_{\text{tab}}$. Due to the construction of the semi-synthetic data, the resulting estimate should be highly biased, as the model is only able to account for the part of the confounding, which is generated via the tabular data.

The \textit{Deep Model} relies on the proposed architecture in Figure \ref{fig:PLRNetwork} and uses the out-of-sample predictions of $\hat{m}(X)$ and $\hat{l}(X)$ generated from the model. As the model utilizes multimodal features $X=(X_{\text{tab}}, X_{\text{txt}}, X_{\text{img}})$, the estimate should be much less biased than the Baseline Model. For our simulation study, we use the RoBERTa Model pretrained on a Twitter Dataset \citep{loureiro2022timelms} as the text model. For the image processing we rely on a VIT Model pretrained on the ImageNet-21k Dataset \citep{rw2019timm}. Both models are implemented in the Hugging Face \texttt{transformers} package \citep{wolf-etal-2020-transformers}. The tabular data is handled by a SAINT model \citep{somepalli2021saint} implemented in the \texttt{pytorch-widedeep} package \citep{Zaurin_pytorch-widedeep_A_flexible_2023}.

The \textit{Embedding Model} also relies on the proposed architecture in Figure \ref{fig:PLRNetwork}, but does not use the generated predictions directly. Instead the generated embedding $H_E$ is used together with the tabular features $X_{\text{tab}}$ as input for a boosting algorithm. Since neural networks are often outperformed by tree based models, such as gradient boosted trees, on tabular data \citep{grinsztajn2022treebased}, the model might perform better on the tabular part of the data, while still accounting for the information contained in the image and text components.

In order to compare the predictive performance of the models, a relative $r^2$-score with respect to the upper bound as described in Section \ref{synth_data} is defined as
\begin{small}
\begin{align*}
    0 \leq r^2(D, \hat{m}) &:= \frac{R^2(D, \hat{m}(X))}{R^2(D, \tilde{m}_0(\tilde{X}))} \leq 1\\
    0 \leq r^2(Y, \hat{l}) &:= \frac{R^2(Y, \hat{l}(X))}{R^2(Y, \tilde{l}_0(\tilde{X}))} \leq 1
\end{align*}
\end{small}
The results of the Baseline Model, Embedding Model and Deep Model are presented in Table \ref{tab:res} with the $r^2$-score values for the confounding functions $\hat{l}$ and $\hat{m}$ as well as the estimated treatment effect $\hat{\theta}$. Notably, the confounding effect is substantial and designed to create a negative bias, wherein higher outcomes $Y$ are associated with lower treatment values $D$. This configuration enables a comprehensive evaluation of bias correction methods, particularly in scenarios where traditional approaches may struggle to account for confounding adequately.
\begin{table}[]
\small
\caption{Results of Simulation Study. Reported: mean ± sd. over five
random train-test splits.}
    \label{tab:res}
\begin{tabular}{lccc}
\toprule
& \textbf{Baseline} & \textbf{Embedding} & \textbf{Deep} \\
\midrule
\small{$r^2(Y, \hat{l}_0)$} & $0.27 \pm 0.02$ & $0.88 \pm 0.02$ & $\bm{0.90 \pm 0.02}$ \\
\small{$r^2(D, \hat{m}_0)$} & $0.29 \pm 0.01$ & $0.89 \pm 0.01$ & $\bm{0.90 \pm 0.01}$ \\
\small{$\hat{\theta}$}      & $-0.32 \pm 0.01$ & $\bm{0.28 \pm 0.01}$ & $\bm{0.28 \pm 0.01}$ \\
\bottomrule
\end{tabular}
\tiny 
Higher = better (best in bold)
\end{table}

\begin{figure}[H]
    \centering
    \centerline{\includegraphics[width=\columnwidth]{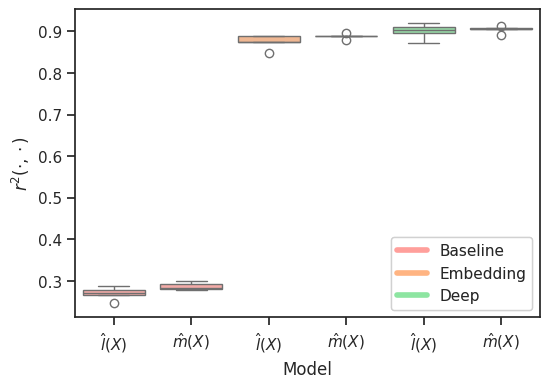}}
    \caption{Boxplots of $r^2$-Scores. As anticipated, the tabular data provides only $30\%$ explanatory power, but the inclusion of unstructured data increases the predictable variance to approximately $90\%$.}
    \label{fig:R2_box}
    
\end{figure}

\begin{figure}[H]
    \centering
    \centerline{\includegraphics[width=\columnwidth]{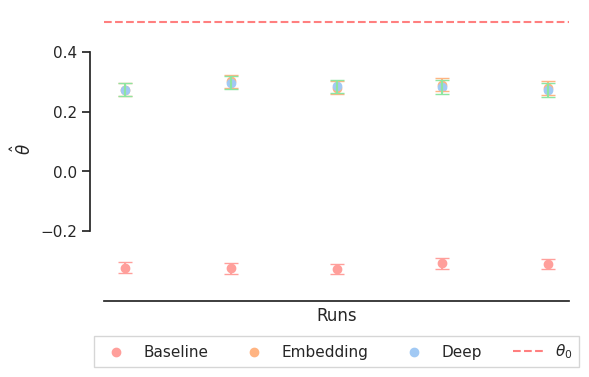}}
    \caption{Boxplots of $\hat{\theta}$. The Embedding Model and Deep Model have similar estimates. This indicates a stable and information-rich embedding $H_E$, which provides a high explanatory contribution independent of the subsequent ML method for predicting $Y$ and $D$ \citep{bengio2014representation}. $\theta_0$ represents the upper bound.}
    \label{fig:theta_hat_box}
\end{figure}

\begin{figure}[H]
    \centering
    \centerline{\includegraphics[width=\columnwidth]{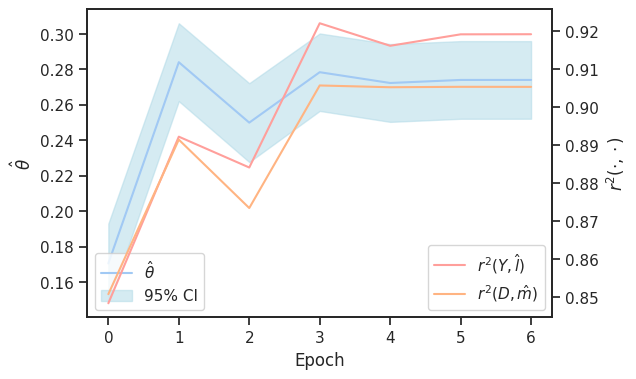}}
    \caption{Continuous estimation of $\theta_0$. Both, the $r^2$-scores and the coefficient estimates gradually converge to stable values over the epochs.}
    \label{fig:dml_deep_estimation_cont}
\end{figure}

The results presented in Table \ref{tab:res} demonstrate the tangible impact of incorporating multimodal information into the estimation process. The Baseline Model, which relies solely on tabular data, exhibits significant bias in estimating the treatment effect ($\hat{\theta} = -0.32 \pm 0.01$), indicative of insufficient control over confounding factors originating from text and image modalities. Figure \ref{fig:R2_box} shows the boxplots of the $r^2$-scores of all three models. This comparative analysis of $r^2$-scores highlights the great predictive performance of the Embedding and Deep Model (approximately $90\%$ of predictable variance), indicating its ability to utilize both structured and unstructured data for more accurate estimation of treatment effects.
Figure \ref{fig:theta_hat_box} shows boxplots of the estimated $\hat{\theta}$ values including the $95\%$ confidence interval.
The Deep Model is able to give a continuous estimation of $\theta_0$ after each training epoch, which is shown in Figure \ref{fig:dml_deep_estimation_cont} including the $95 \%$ confidence interval as well. As can be seen, the estimation $\hat{\theta}$ gets closer to the true value $\theta_0$ during the training process. The observed trends in the $r^2$-scores ($r^2(Y, \hat{l})$ and $r^2(D, \hat{m})$) and in the coefficient estimate ($\hat{\theta}$) provide valuable insights into the relationship between the prediction performance and the bias correction of the causal estimate.


\section{Conclusion}
In this article we extend the double machine learning framework to allow for tabular data, text, and images as confounding variables. This makes it possible to provide valid inference of the causal target parameter in this setting. Given the increasing availability of multimodal data, we consider our paper as an important innovation in the field of causal machine learning. Incorporating information from images or text can help to improve results from causal studies, either by capturing otherwise unmeasured confounding or by improving the precision of statistical estimators. To validate our approach, we set up a new framework to generate semi-synthetic data with multimodal data that addresses inherent challenges in this type of simulation studies. The proposed method might be of independent interest to the scientific community. Our semi-synthetic numerical experiments demonstrated the model's ability to incorporate the confounding that is contained in the text and image data. The performance substantially improved as compared to a benchmark model that does not account for this information. We acknowledge that in our simulations our model does not consistently estimate the true causal parameter. This, however, is rather the consequence of the inherent limitations to simulated text and image data for causal analysis rather than a shortcoming of our neural network architecture, that is based on state-of-the-art approaches.
To the best of our knowledge, we are not aware of any other approach to handle multimodal data for causal inference. Our approach was tailored to the partially linear regression model but is generally applicable to any kind of causal model that fits in the double machine learning framework, for example nonparametric models to estimate treatment effects or causal models that explicitly address variation over time, such as panel data or difference-in-differences models. In these cases, the architecture of our neural network would have to be adjusted to the definitions of the nuisance components in these models. Moreover, our approach could also be extended to other kinds of unstructured data, like graphs, networks, audio or video data.

\section*{Impact Statement}
This paper presents work whose goal is to advance the fields of Machine Learning and Causal Inference. There are many potential societal consequences of our work, none of which we feel must be specifically highlighted here.

\bibliography{dml_deep_proceeding}
\bibliographystyle{icml2024}

\newpage
\appendix
\onecolumn
\section{Appendix}
\subsection{Definitions}
Let $V$ be a random variable and $V_1,\dots, V_n$ iid. realizations of $V$.\\
Define 
\begin{align*}
    \mathbb{E}_n [V]:= \frac{1}{n}\sum_{i=1}{n}V_i
\end{align*}
and the correspondingly
\begin{align*}
    \|V\|_{P,2} &:= \mathbb{E}[V^2]\\
    \|V\|_{P_n,2} &:= \mathbb{E}_n[V^2].
\end{align*}

Further, define
\begin{align*}
    R^2(V, \hat{V}) = 1 - \frac{\sum_{i=1}^n (V_i - \hat{V}_i)^2}{\sum_{i=1}^n (V_i - \mathbb{E}_n[V])^2}.
\end{align*}
for two random variables $V$ and $\hat{V}$.

\subsection{Semi-Synthetic Dataset} \label{app_semisynthetic}
This subsection includes further information regarding the semi-synthetic dataset.
Due to the distribution of the noise and centering of the confounding components it should hold
\begin{align*}
    \mathbb{E}[Y] &= \mathbb{E}[D] = 0\\
    \text{Var}(Y) &= \text{Var}(D) = 3.
\end{align*}
Descriptives regarding the treatment and outcome variable. 
\begin{center}
\begin{tabular}{lrr}
\toprule
 & Y & D \\
\midrule
count & 50000 & 50000 \\
mean & 0.006948 & -0.000755 \\
std & 1.734030 & 1.738060 \\
min & -6.233601 & -6.461094 \\
25\%-quantile & -1.184978 & -1.208271 \\
50\%-quantile & 0.014107 & -0.000240 \\
75\%-quantile & 1.190921 & 1.193110 \\
max & 6.169482 & 6.260997 \\
\bottomrule
\end{tabular}
\end{center}
The oracle root mean squared errors for the nuisance components are
\begin{align*}
    \|D - \tilde{m}_0(\tilde{X})\|_{P_n,2} &= 0.9965 \\
    \|Y - \tilde{l}_0(\tilde{X})\|_{P_n,2} &= 1.1177.
\end{align*}


\end{document}